\documentclass[pdflatex,sn-mathphys, iicol]{sn-jnl}

\jyear{2022}%

\theoremstyle{thmstyleone}%
\theoremstyle{thmstyletwo}%

\theoremstyle{thmstylethree}%

\begin{document}

\title[Article Title]{Comment on "Black Box Prediction Methods in Sports Medicine Deserve a Red Card for Reckless Practice: A Change of Tactics is Needed to Advance Athlete Care"}


\author*[1]{\fnm{Jakim} \sur{Berndsen}}\email{jberndsen@kitmanlabs.com}

\author[1]{\fnm{Derek} \sur{McHugh}}\email{derek@kitmanlabs.com}
\equalcont{These authors contributed equally to this work.}


\affil*[1]{\orgname{Kitman Labs}, \orgaddress{\city{Dublin}, \country{Ireland}}}

\maketitle

Dear Editor,
\\

We read with interest the article by Bullock et. al. \cite{bullock2022black} on the applicability of Machine Learning (ML) approaches in sports medicine settings. Overall, we agree with the central thrust of this paper—without sufficient interpretability, ML models are not useful in the sports injury domain. As the authors correctly state, black box models can cause problems. If we cannot establish how a model is making its decisions, and the predictive power is low, there is no value to using these models.

However, we feel there are areas of research within ML that have been overlooked which strongly temper the conclusions that the authors make. In the following,  we look at how these are being addressed by the ML community.

\section{Claim: ML models are black boxes}
This assertion depends on the ML model being deployed. While deep learning and boosting approaches may lack interpretability, simpler models, such as decision trees, can be considered machine learning and are highly interpretable. Newer modelling techniques, such as Explainable Boosting Machines \cite{nori2019interpretml}, produce great results with interpretability as a cornerstone of their design.

Explainable AI techniques can also be used to open the black box of more complex models. While solutions are not yet perfect, advancements such as LIME \cite{ribeiro2016should} and Shap \cite{lundberg2017unified} allow us to understand the inner workings of complex, multivariate models and can help us unravel our algorithms in an interpretable manner.

\section{Claim: Existing evidence cannot be used in ML models}
This statement depends on how a ML model is built. While it is true that neural network architectures can achieve state of the art performances on raw data, there are many applications where this approach is inappropriate. The decision is typically based on the goals of your model, e.g., whether you are looking for pure performance, or wish to learn something about the underlying truth in your data. Sometimes, this decision can be made for you, based on the availability and quality of your data, which is often messy and incomplete in sporting environments.

Domain knowledge can be incorporated into ML solutions in a variety of ways:
\begin{itemize}
    \item Informing and defining research questions after data exploration
    \item Feature selection - identifying and using causal features
    \item Feature engineering - creating new features based on prior knowledge of injuries, or existing research
\end{itemize}

\section{Claim: Models cannot provide insight into 1) what interventions should be implemented and 2) the magnitude of those interventions}
A further field of ML, beyond predictive modelling, is that of Recommender Systems (RS) \cite{10.5555/1941884}. Typically, RS help users find items they may like, such as films on Netflix, or products on Amazon. However, RS techniques have been used in both sporting and medical domains.

In a sporting context, RS have been used to help athletes train and plan for their events in sports such as running \cite{berndsen2019pace,smyth2021recommendations}. In a medical setting, RS can be used for diet prescription \cite{agapito2017dietos} or to aid sleep \cite{pandey2020personalized}. In all of these cases, RS techniques are  used to identify the type and magnitude of intervention needed. In fact, early research has looked at injury mitigation using RS in recreational marathon runners \cite{feely2021case}.

Currently, a holistic medical solution is not yet available, but significant progress is being made applying RS to sports injuries. An RS solution need not replace the expertise of a practitioner, and can instead act as a decision-support tool in a \emph{human-in-the-loop} \cite{holzinger2016interactive} system.

\section{Claim: Robust Model Validation is critical for ML solutions to be useful}
We agree with the authors that robust validation of models is needed. Without reliable and correctly implemented validation procedures, ML models should not be trusted. However, we believe that it is practically impossible to have a global model for injuries in sport. Sports injury models are dependent on the context in which they are trained, and will vary depending on team practice. For example, a team may have an excellent hamstring rehabilitation programme, and the model will therefore show a negative relationship between injury risk and previous hamstring injuries. In other teams—or on a global level—this is unlikely to be the case, as we know previous injuries tend to be indicative of future injuries. The model with the negative relationship is therefore only relevant for the team it was trained on but this is not a problem as long as it is adding value. Injury risk may be governed by thousands of local models, which is fine so long as each is adding useful information.

Advantages of ML models
With the above in mind, it’s important to highlight a few of the advantages that ML models can provide over more simplistic approaches that the authors are advocating for. ML models:
\begin{itemize}
    \item Are not constrained by the same assumptions as simple models, e.g., linear relationships or independence
    \item Can handle a large number of features to produce a multifaceted model of injury risk
    \item Are less sensitive to noise than simpler models (Can easily handle missing values which often occur in sports data, e.g., a device has malfunctioned, or no time to conduct a particular test)
    \item Offer many solutions to address the imbalance in injury data, whereas simple models often only predict the majority class in such situations, which teaches us nothing about injury
\end{itemize}

\section{Conclusion}
Overall we agree with many of the sentiments of the opinion piece. However, there is an immense amount of work being done within the ML (and sport science) community that addresses precisely the issues and concerns the paper has about ML approaches in the sports medicine field. While further progress is needed, and the current solutions may not suit everybody, injury risk analytics can offer transparency and practical insights to support practitioners in their quest to reduce injury rates. We think there is a clear case for the red card to be rescinded upon appeal, and propose it might have been a yellow at worst.


\bibliography{sn-bibliography}


\begin{thebibliography}{11}
\ifx \bisbn   \undefined \def \bisbn  #1{ISBN #1}\fi
\ifx \binits  \undefined \def \binits#1{#1}\fi
\ifx \bauthor  \undefined \def \bauthor#1{#1}\fi
\ifx \batitle  \undefined \def \batitle#1{#1}\fi
\ifx \bjtitle  \undefined \def \bjtitle#1{#1}\fi
\ifx \bvolume  \undefined \def \bvolume#1{\textbf{#1}}\fi
\ifx \byear  \undefined \def \byear#1{#1}\fi
\ifx \bissue  \undefined \def \bissue#1{#1}\fi
\ifx \bfpage  \undefined \def \bfpage#1{#1}\fi
\ifx \blpage  \undefined \def \blpage #1{#1}\fi
\ifx \burl  \undefined \def \burl#1{\textsf{#1}}\fi
\ifx \doiurl  \undefined \def \doiurl#1{\url{https://doi.org/#1}}\fi
\ifx \betal  \undefined \def \betal{\textit{et al.}}\fi
\ifx \binstitute  \undefined \def \binstitute#1{#1}\fi
\ifx \binstitutionaled  \undefined \def \binstitutionaled#1{#1}\fi
\ifx \bctitle  \undefined \def \bctitle#1{#1}\fi
\ifx \beditor  \undefined \def \beditor#1{#1}\fi
\ifx \bpublisher  \undefined \def \bpublisher#1{#1}\fi
\ifx \bbtitle  \undefined \def \bbtitle#1{#1}\fi
\ifx \bedition  \undefined \def \bedition#1{#1}\fi
\ifx \bseriesno  \undefined \def \bseriesno#1{#1}\fi
\ifx \blocation  \undefined \def \blocation#1{#1}\fi
\ifx \bsertitle  \undefined \def \bsertitle#1{#1}\fi
\ifx \bsnm \undefined \def \bsnm#1{#1}\fi
\ifx \bsuffix \undefined \def \bsuffix#1{#1}\fi
\ifx \bparticle \undefined \def \bparticle#1{#1}\fi
\ifx \barticle \undefined \def \barticle#1{#1}\fi
\bibcommenthead
\ifx \bconfdate \undefined \def \bconfdate #1{#1}\fi
\ifx \botherref \undefined \def \botherref #1{#1}\fi
\ifx \url \undefined \def \url#1{\textsf{#1}}\fi
\ifx \bchapter \undefined \def \bchapter#1{#1}\fi
\ifx \bbook \undefined \def \bbook#1{#1}\fi
\ifx \bcomment \undefined \def \bcomment#1{#1}\fi
\ifx \oauthor \undefined \def \oauthor#1{#1}\fi
\ifx \citeauthoryear \undefined \def \citeauthoryear#1{#1}\fi
\ifx \endbibitem  \undefined \def \endbibitem {}\fi
\ifx \bconflocation  \undefined \def \bconflocation#1{#1}\fi
\ifx \arxivurl  \undefined \def \arxivurl#1{\textsf{#1}}\fi
\csname PreBibitemsHook\endcsname

\bibitem{bullock2022black}
\begin{botherref}
\oauthor{\bsnm{Bullock}, \binits{G.S.}},
\oauthor{\bsnm{Hughes}, \binits{T.}},
\oauthor{\bsnm{Arundale}, \binits{A.H.}},
\oauthor{\bsnm{Ward}, \binits{P.}},
\oauthor{\bsnm{Collins}, \binits{G.S.}},
\oauthor{\bsnm{Kluzek}, \binits{S.}}:
Black box prediction methods in sports medicine deserve a red card for reckless
  practice: A change of tactics is needed to advance athlete care.
Sports Medicine,
1--7
(2022)
\end{botherref}
\endbibitem

\bibitem{nori2019interpretml}
\begin{botherref}
\oauthor{\bsnm{Nori}, \binits{H.}},
\oauthor{\bsnm{Jenkins}, \binits{S.}},
\oauthor{\bsnm{Koch}, \binits{P.}},
\oauthor{\bsnm{Caruana}, \binits{R.}}:
Interpretml: A unified framework for machine learning interpretability.
arXiv preprint arXiv:1909.09223
(2019)
\end{botherref}
\endbibitem

\bibitem{ribeiro2016should}
\begin{bchapter}
\bauthor{\bsnm{Ribeiro}, \binits{M.T.}},
\bauthor{\bsnm{Singh}, \binits{S.}},
\bauthor{\bsnm{Guestrin}, \binits{C.}}:
\bctitle{" why should i trust you?" explaining the predictions of any
  classifier}.
In: \bbtitle{Proceedings of the 22nd ACM SIGKDD International Conference on
  Knowledge Discovery and Data Mining},
pp. \bfpage{1135}--\blpage{1144}
(\byear{2016})
\end{bchapter}
\endbibitem

\bibitem{lundberg2017unified}
\begin{botherref}
\oauthor{\bsnm{Lundberg}, \binits{S.M.}},
\oauthor{\bsnm{Lee}, \binits{S.-I.}}:
A unified approach to interpreting model predictions.
Advances in neural information processing systems
\textbf{30}
(2017)
\end{botherref}
\endbibitem

\bibitem{10.5555/1941884}
\begin{bbook}
\bauthor{\bsnm{Ricci}, \binits{F.}},
\bauthor{\bsnm{Rokach}, \binits{L.}},
\bauthor{\bsnm{Shapira}, \binits{B.}},
\bauthor{\bsnm{Kantor}, \binits{P.B.}}:
\bbtitle{Recommender Systems Handbook},
\bedition{1st} edn.
\bpublisher{Springer},
\blocation{Berlin, Heidelberg}
(\byear{2010})
\end{bbook}
\endbibitem

\bibitem{berndsen2019pace}
\begin{bchapter}
\bauthor{\bsnm{Berndsen}, \binits{J.}},
\bauthor{\bsnm{Smyth}, \binits{B.}},
\bauthor{\bsnm{Lawlor}, \binits{A.}}:
\bctitle{Pace my race: recommendations for marathon running}.
In: \bbtitle{Proceedings of the 13th ACM Conference on Recommender Systems},
pp. \bfpage{246}--\blpage{250}
(\byear{2019})
\end{bchapter}
\endbibitem

\bibitem{smyth2021recommendations}
\begin{botherref}
\oauthor{\bsnm{Smyth}, \binits{B.}},
\oauthor{\bsnm{Lawlor}, \binits{A.}},
\oauthor{\bsnm{Berndsen}, \binits{J.}},
\oauthor{\bsnm{Feely}, \binits{C.}}:
Recommendations for marathon runners: on the application of recommender systems
  and machine learning to support recreational marathon runners.
User Modeling and User-Adapted Interaction,
1--52
(2021)
\end{botherref}
\endbibitem

\bibitem{agapito2017dietos}
\begin{bchapter}
\bauthor{\bsnm{Agapito}, \binits{G.}},
\bauthor{\bsnm{Simeoni}, \binits{M.}},
\bauthor{\bsnm{Calabrese}, \binits{B.}},
\bauthor{\bsnm{Guzzi}, \binits{P.H.}},
\bauthor{\bsnm{Fuiano}, \binits{G.}},
\bauthor{\bsnm{Cannataro}, \binits{M.}}:
\bctitle{Dietos: A recommender system for health profiling and diet management
  in chronic diseases.}
In: \bbtitle{HealthRecSys@ RecSys},
pp. \bfpage{32}--\blpage{35}
(\byear{2017})
\end{bchapter}
\endbibitem

\bibitem{pandey2020personalized}
\begin{bchapter}
\bauthor{\bsnm{Pandey}, \binits{V.}},
\bauthor{\bsnm{Upadhyay}, \binits{D.D.}},
\bauthor{\bsnm{Nag}, \binits{N.}},
\bauthor{\bsnm{Jain}, \binits{R.C.}}:
\bctitle{Personalized user modelling for context-aware lifestyle
  recommendations to improve sleep.}
In: \bbtitle{HealthRecSys@ RecSys},
pp. \bfpage{8}--\blpage{14}
(\byear{2020})
\end{bchapter}
\endbibitem

\bibitem{feely2021case}
\begin{bchapter}
\bauthor{\bsnm{Feely}, \binits{C.}},
\bauthor{\bsnm{Caulfield}, \binits{B.}},
\bauthor{\bsnm{Lawlor}, \binits{A.}},
\bauthor{\bsnm{Smyth}, \binits{B.}}:
\bctitle{A case-based reasoning approach to predicting and explaining running
  related injuries}.
In: \bbtitle{International Conference on Case-Based Reasoning},
pp. \bfpage{79}--\blpage{93}
(\byear{2021}).
\bcomment{Springer}
\end{bchapter}
\endbibitem

\bibitem{holzinger2016interactive}
\begin{barticle}
\bauthor{\bsnm{Holzinger}, \binits{A.}}:
\batitle{Interactive machine learning for health informatics: when do we need
  the human-in-the-loop?}
\bjtitle{Brain Informatics}
\bvolume{3}(\bissue{2}),
\bfpage{119}--\blpage{131}
(\byear{2016})
\end{barticle}
\endbibitem

\end{thebibliography}


\end{document}